# Estimation of 3D Human Pose Using Prior Knowledge

Shu Chen, Lei Zhang and Beiji Zou

*Abstract*—Estimating three-dimensional (3D) human poses from the positions of two-dimensional (2D) joints has shown promising results. However, using 2D joint coordinates as input loses more information than image-based approaches and results in ambiguity. In order to overcome this problem, we combine bone length and camera parameters with 2D joint coordinates for input. This combination is more discriminative than the 2D joint coordinates in that it can improve the accuracy of the model's prediction depth and alleviate the ambiguity that comes from projecting 3D coordinates into 2D space. Furthermore, we introduce direction constraints which can better measure the difference between the ground truth and the output of the proposed model. The experimental results on the Human3.6M show that the method performed better than other state-of-the-art 3D human pose estimation approaches. The code is available at https://github.com/XTU-PR-LAB/ExtraPose

## I. Introduction

Human pose estimation is a popular research field in computer vision, and there are many applicable scenarios for this research, such as games, animation, action recognition, motion capture and augmented reality. Many approaches in this field have successfully inferred three-dimensional (3D) human poses from images or videos [2, 3, 4]. Other methods predict 3D human poses from two-dimensional (2D) human joints which are produced by a 2D joint detector or by projecting the 3D ground truth onto the image plane [5, 6, 7]. These approaches do not use camera parameters nor the information about human bones.

Camera parameters play an important role in predicting 3D human poses, whether projecting 3D ground truth into 2D space or transforming 3D poses from world coordinates to a camera coordinate system. Bastian *et al*. [8] use a linear model to predict a 6D vector, in which the camera parameters are included for projecting, because six variables can define a weak camera model. Their network generalizes well to unknown data; even strong deformations and unusual camera poses can be reconstructed. Jogendra Nath Kundu *et al*. [14] get the camera extrinsic by training an encoder model, whereas a fixed perspective camera projective camera projection is applied to obtain the final 2D pose representation. Ikhsanul Habibie *et al*. [15] use a multi-layer perceptron to infer the principal coordinates and the focal length parameters of a weak perspective camera model from the given input image. Rongchang Xie *et al*. [9] learn a pre-trained feature fusion model on a large-scale multi-camera data, to find the corresponding locations between different cameras and maximize the generalization ability of the model. This method solves the limb occlusion problem because a joint occluded in one view could be visible in other views. Wei *et al*. [16] alleviate the effects of viewpoint diversity by using a new view-invariant module, and achieve significant improvement. Zhao *et al*. [18] focus on fusing image information from multiple cameras in different views.

Iqbal *et al*. [11] train the network to predict a 2.5D pose representation from which the 3D pose can be reconstructed in a fully differentiable way. The 2.5D representation consists of 2D joint position and the relative depth, and it has several key features to exploit multi-view information and devise loss function for weakly-supervised training. Wang *et al*. [12] proposed the Pairwise Ranking Convolutional Neural Network to estimate the depth ranking that is presented by using a pairwise ranking matrix. With the knowledge of depth rankings between adjacent joints, 2D joint locations, and limb length priors, the 3D skeleton is almost determined. Sharma *et al*. [13] tackle the ambiguity in the 2D-to-3D mapping by training a Deep Conditional Variational Autoencoder (CAVE) model to generate 3D-pose samples conditioned on the 2D pose, that are scored and weighted-average using joint-ordinal relations, which are regressed together with the 2D pose.

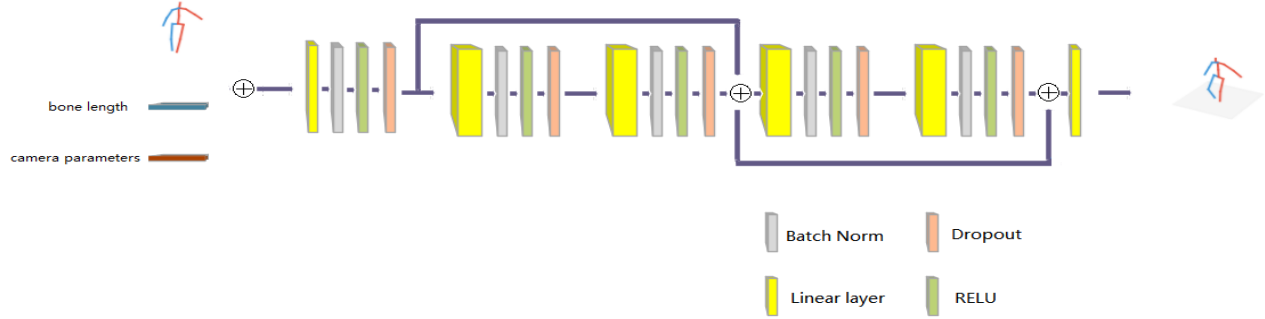

Fig. 1. The architecture of the lifting model to regress 3D human pose. The input is the combination of the 2D joint position, bone length and camera intrinsic parameters and the output is the predicted 3D human pose. The loss include two parts: the L2 loss between the ground truth and the 3D prediction and the direction loss.

These methods use camera parameters or bone length in different ways, but we argue that the combination of camera parameters and bone length could have a good impact on the accuracy of 3D single human pose estimation.

## II. Approach

Our approach infers 3D human pose from 2D human joints using bone length and camera parameters. As shown in Figure 1, we lift 3D human joints not just from 2D joints' positions, but the combination of three parts which could help our network to achieve better results in inferring 3D human poses.

Recovering 3D poses purely from 2D joints' locations is an ill-posed problem, as multiple 3D human poses have the same 2D projection. Moreover, the information contained in the 2D data is limited, and the predicted results are not often particularly accurate. The use of prior knowledge could alleviate these problems. The bone length and camera parameters are used as prior knowledge in this work. Together with the 2D pose, they enrich the information of the input data, so that the focus of the neural network model is not limited to the plane coordinates of the joint points, but also the geometric information of the human body and external spatial information.

### A. Combination of camera and bone length

Using the focal length and focus of the camera, we can build a weak camera model. According to the principle of small hole imaging, we can use the weak camera model to project a 3D human body posture onto a 2D plane. At the opposite stage, due to the introduction of camera parameters and bone length parameters, the model's prediction of depth can be affected.

In the camera's perspective, the 3D coordinates are returned from the 2d joint points. Then the possible solutions of its $z$ coordinate are on a straight line between the focal point and the joint point. The introduction of bone length reduces the number of possible solutions, so we can construct a circle with the joint point as the center and the bone length as the radius. The straight line formed by the focal point and the joint point will pass through the circle, and the $z$ coordinate will be on the intersection of the circle and the straight line. The combination of bone length and camera parameters alleviates the ambiguity of 3D human poses lifted from the 2D joints' locations, to a certain extent, and improves the accuracy of model prediction results.

### B. Camera parameters normalization

Since the focal length and focus parameters in the camera are far from the 2D coordinates of the human body joints, they need to be normalized. For the focal point, the original locations range between [0, $w$] is normalized to between [-1, 1] which preserves the aspect ratio. The normalization of the focal length is achieved by first dividing by the width of the camera resolution and then multiplying by two. They are normalized as

$$focus = \begin{bmatrix} 2 \cdot \frac{x}{w} \\ 2 \cdot \frac{y}{w} \end{bmatrix} - \begin{bmatrix} 1 \\ \frac{h}{w} \end{bmatrix}, \quad (1)$$

$$focal\ length = 2 \cdot \frac{focal\ length}{w}, \quad (2)$$

where $h$ and $w$ are the height and width of resolution, respectively.

### C. Bone length

The number of 2D human joint points detected by the hourglass model is 16, so the calculated number of bone lengths is 15. In order to avoid excessive span between data, the calculation of bone length is standardized in the 2D joint point coordinates. After that, the formula is defined as follows:

$$bone\ length = \sqrt{(x_c - x_p)^2 + (y_c - y_p)^2 + (z_c - z_p)^2}, \quad (3)$$

where $c$ represents the child joint node, and $p$ refers to the relative parent joint node.

TABLE I
EVALUATION ON HUMAN 3.6M WITH PROTOCOL-1. *GT* REPRESENTS THE GROUND TRUTH DATASET

| Protocol-1 | Direct | Disc | Eat | Greet | Phone | Photo | Pose | Purch | Sit | SitD | Smoke | Wait | WalkD | Walk | WalkT | Avg |
|---|---|---|---|---|---|---|---|---|---|---|---|---|---|---|---|---|
| Zhou[21] | 54.8 | 60.7 | 58.2 | 71.4 | 62.0 | 65.5 | 53.8 | 55.6 | 75.2 | 111 | 64.2 | 66.1 | 51.4 | 63.2 | 55.3 | 64.9 |
| Matinze[5] | 50.3 | 54.4 | 56.9 | 58.0 | 66.4 | 76.0 | 53.4 | 55.4 | 72.9 | 91.3 | 60.7 | 58.4 | 63.0 | 48.8 | 52.0 | 61.2 |
| Yang[22] | 51.5 | 58.9 | 50.4 | 57.0 | 62.1 | 65.4 | 49.8 | 52.7 | 69.2 | 85.2 | 57.4 | 58.4 | 43.6 | 60.1 | 47.7 | 58.6 |
| Pavlakos[23] | 48.5 | 54.4 | 54.4 | 52.0 | 59.4 | 65.3 | 49.9 | 52.9 | 65.8 | 71.1 | 56.6 | 52.9 | 60.9 | 44.7 | 47.8 | 56.2 |
| Zhao[6] | 47.3 | 60.7 | 51.4 | 60.5 | 61.1 | 49.9 | 47.3 | 68.1 | 86.2 | 55.0 | 67.8 | 61.0 | 42.1 | 60.6 | 45.3 | 57.6 |
| Pavllo[4] | 47.1 | 50.6 | 49.0 | 51.8 | 53.6 | 61.4 | 49.4 | 47.4 | 59.3 | 67.4 | 52.4 | 49.5 | 55.3 | 39.5 | 42.7 | 51.8 |
| Xu[19] | 40.6 | 47.1 | 45.7 | 46.6 | **50.7** | 63.1 | 45.0 | 47.7 | **56.3** | **63.9** | **49.4** | 46.5 | 51.9 | **38.1** | 42.3 | 49.2 |
| ours | **39.9** | **45.8** | **43.7** | **45.3** | 53.1 | **59.8** | **43.7** | **44.7** | 59.3 | 70.7 | 49.9 | **45.6** | **49.5** | 38.2 | **43.3** | **48.8** |
| ours(gt) | 35.8 | 41.9 | 40.0 | 40.6 | 44.7 | 51.4 | 42.8 | 39.1 | 53.2 | 55.3 | 42.8 | 42.9 | 43.5 | 33.5 | 36.5 | 42.9 |

TABLE II
EVALUATION ON HUMAN 3.6M WITH PROTOCOL-2. *GT* REPRESENTS THE GROUND TRUTH DATASET.

| Protocol-2 | Direct | Disc | Eat | Greet | Phone | Photo | Pose | Purch | Sit | SitD | Smoke | Wait | WalkD | Walk | WalkT | Avg |
|---|---|---|---|---|---|---|---|---|---|---|---|---|---|---|---|---|
| Bogo[17] | 62.0 | 60.2 | 67.8 | 79.5 | 92.1 | 77.0 | 73.0 | 75.3 | 100 | 137 | 83.4 | 77.3 | 86.8 | 79.7 | 87.7 | 82.3 |
| Matinze[5] | 39.1 | 42.2 | 44.6 | 46.5 | 50.0 | 54.1 | 40.3 | 39.1 | 55.7 | 65.5 | 48.7 | 44.1 | 48.7 | 37.8 | 42.5 | 46.6 |
| Lee[24] | 38.0 | 39.3 | 46.3 | 44.4 | 49.0 | 55.1 | 40.2 | 41.1 | 53.2 | 68.9 | 51.0 | 39.1 | 33.9 | 56.4 | 38.5 | 46.2 |
| Fang[20] | 38.2 | 41.7 | 43.7 | 44.9 | 48.5 | 55.3 | 40.2 | 38.2 | 54.5 | 64.4 | 47.2 | 44.3 | 47.3 | 36.7 | 41.7 | 45.7 |
| Pavlakos[23] | 34.7 | 39.8 | 41.8 | 38.6 | 42.5 | 47.5 | 38.0 | 36.6 | 50.7 | 56.8 | 42.6 | 39.6 | 43.9 | 32.1 | 36.5 | 41.8 |
| Pavllo[4] | 36.0 | 38.7 | 38.0 | 41.7 | 40.1 | 45.9 | 37.1 | 35.4 | 46.8 | 53.4 | 41.4 | 36.9 | 43.1 | 30.3 | 34.8 | 40.0 |
| Xu[19] | 33.6 | 37.4 | 37.0 | **37.6** | **39.2** | 46.4 | **34.3** | 35.4 | **45.1** | 52.1 | **40.1** | **35.5** | 42.1 | **29.8** | 35.3 | **38.9** |
| ours | **33.6** | **36.7** | **35.5** | 38.5 | 42.9 | **45.7** | 35.8 | **32.9** | 45.8 | **51.7** | 41.5 | 35.9 | **39.8** | 31.2 | 37.2 | 39.0 |
| ours(gt) | 23.3 | 26.8 | 23.7 | 26.5 | 28.2 | 33.9 | 27.0 | 23.1 | 33.2 | 35.1 | 28.3 | 26.1 | 28.1 | 22.1 | 25.2 | 27.4 |

## D. Direction Loss

In the previous method, bone length loss is added to the final prediction result to constrain the human bone length, and the loss reduces the error caused by the single use of the mean square error as the loss function between the predicted value and the ground truth. The reason the bone length can be used as the constraint is that the bone length of the human body is rigid

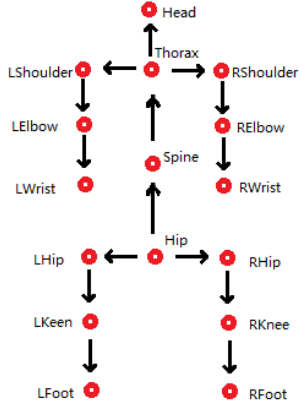

Fig. 2. The orientation information between the parent and child joint point.

and does not change during the movement of the human body.

The bones in the 3D space are unique, and the orientation of the bones of each part of the body also must be unique. We subtract the coordinates of the current joint point from the coordinates of its parent node to get the current bone orientation. The current bone orientation is used as the measuring criterion between the model's predicted value and the ground truth. The direction loss is defined as

$$direction = (x_d, y_d, z_d),  \quad (4)$$

$$x_d = x_c - x_p, \quad (5)$$

$$y_d = y_c - y_p, \quad (6)$$

$$z_d = z_c - z_p, \quad (7)$$

$$loss = \frac{1}{m}\sum_{1}^{m}(direct_{pred} - direct_{gt})^2. \quad (8)$$

The calculated result is a vector, where $d$ denotes the current element which belongs to the direction vector; $c$ and $p$ respectively represent the current joint point and its parent joint point.

The direction information between the parent and child nodes is shown in Figure 2.

## III. EXPERIMENTS

The used datasets and protocols are explained as follows:

### A. Implementation Details

We pre-calculate the bone length of the human body based on the ground truth and standardize the acquired camera external parameters and bone length. In the data preparation stage, the bone length and camera parameters append to the end of the 2D joint point data. To ensure the reliability of the data, we trained both models for 400 epochs, and the training time of each batch did not exceed 17 milliseconds at most. We used an Nvidia 2080ti graphics card to complete the entire training process.

At the end of training, we not only use the mean square error loss between the ground truth and the predicted value, but also introduce the direction loss to further improve the accuracy of the model. By changing the weights of the mean square error loss and the direction loss, we found that when the weights of the two losses were each set to 0.5, the effect of the model was significantly improved. If the weight of the mean square error loss is too large, the constraint effect in the direction will be lost. Meanwhile, if the weight of the direction loss is set too large, then the training result is much worse than without the direction loss.

*B. 3D dataset*

The employed dataset in the experiments is the Human3.6M [10], which is one of the largest datasets for 3D human pose estimation with 3.6 million images. The Human3.6M includes 17 scenarios, such as directions, smoking and sitting, from four cameras views. We mainly used this dataset for training and evaluating the model.

We split the dataset into two sets: the training set and the testing set. Subjects 1, 5, 6, 7 and 8 belong to the training set and were used for training, Subjects 9 and 11 in the testing set and were used for evaluating. Each subject has fifteen actions (Directions, Discussion, Eating, Greeting, Phoning, Photo, Posing, Purchases, Sitting, Sitting Down, Smoking, Waiting, Walk Dog, Walking, Walk Together).

*C. 2D datasets*

There were two 2D human joint position datasets used in our experiments. The first was the 2d joint detections, which was obtained by the state-of-the-art stacked hourglass network [1]. The stacked hourglass was pretrained on the MPII dataset and then fine-tuned on the Human3.6M dataset.

The second was the ground truth dataset. The samples were produced by projecting 3D human joint positions into 2D coordinate space utilizing camera parameters.

*D. Evaluation protocols*

The standard protocols are mean per joint position error (MPJPE) and Procrustes aligned mean per joint position error (P-MPJPE). MPJPE calculates the Euclidean distance between the predicted 3D human pose and ground truth to evaluate the performance of the model. The formula is defined as

$$MPJPE = \frac{\Sigma\left(\sqrt[2]{(x_i - x_{gt})^2 + (y_i - y_{gt})^2 + (z_i - z_{gt})^2}\right)}{n}, \quad (9)$$

where $n$ is the total number of the human joints; $gt$ is the ground truth, and $(x, y, z)$ is the human joint position in 3D space.

P-MPJPE performs rigid transformation on the output to align with the ground truth, and then calculates MPJPE.

*E. Quantitative Results*

We compared the performance of our approach with several state-of-the-art algorithms on Human3.6M, as shown in Tables I and II. The results show the 3D human pose obtained by out method is more accurate. From the tables, we can find that the accuracy of certain actions is significantly better than the algorithm proposed by Xu *et al.* [19].

TABLE IV
ABLATION

|  | P1 | P2 |
|---|---|---|
| *Mdn[7] (gt)* | 36.5 | 30.1 |
| *+ camera and bone length* | 35.9 | 26.5 |
| *Mdn[7] (sh)* | 52.7 | 42.6 |
| *+ camera and bone length* | 48.0 | 37.6 |

In order to verify that our method is valid on other models, we used Mdn[7] for experiments. The results are shown in Table IV. Compared with simply taking 2D joint points as input, after adding knowledge prior, the result of the 3D human posture prediction has been significantly improved.

*F. Ablation Study*

We conducted ablation experiments to verify the effectiveness of the method. Based on the Matinze's model [5], we evaluated the effectiveness of the input combined with bone length and camera parameters and the direction loss on the ground truth data set and the hourglass model detection data set. The bone length, camera parameters, and direction loss were added separately, and the results are shown in Table III.

From Table III, we find that after adding the bone length and camera parameters separately, the accuracy of the model's prediction of the 3D human pose improved to varying degrees. The accuracy of the results is most improved when the bone length and camera parameters were added at the same time. At the end, directional loss was introduced, and the best results were achieved on both the ground truth data set and the hourglass data set.

IV. CONCLUSION

Our method has exerted very good effect on the neural network model of 3D human posture estimation based on 2D joint coordinate points. Our experiment verifies that with camera parameters and bone length respectively, the performance can be improved to a certain extent, and with the combination of two parameters, the experimental results can be greatly improved. To a certain extent, our method alleviate the ambiguity of 3D human pose lifting from the corresponding 2D

TABLE III
ABLATION

|  | P1 | P2 |
|---|---|---|
| *Matinze [5](gt)* | 42.98 | 33.89 |
| *+ camera* | 39.89 | 29.81 |
| *+ bone length* | 38.13 | 31.09 |
| *+ camera and bone length* | 35.95 | 27.61 |
| *+direction loss* | 35.87 | 27.43 |
| *Matinze(sh)* | 61.23 | 46.64 |
| *+ camera* | 56.49 | 43.35 |
| *+ bone length* | 50.78 | 41.06 |
| *+ camera and bone length* | 49.00 | 39.12 |
| *Direction* | 48.86 | 39.03 |

joints locations.